\documentclass[conference]{IEEEtran}
\IEEEoverridecommandlockouts
\usepackage{cite}
\usepackage{amsmath,amssymb,amsfonts}
\usepackage{algorithmic}
\usepackage{graphicx}
\usepackage{textcomp}
\usepackage{xcolor}
\usepackage{url}

\def\BibTeX{{\rm B\kern-.05em{\sc i\kern-.025em b}\kern-.08em
    T\kern-.1667em\lower.7ex\hbox{E}\kern-.125emX}}
\begin{document}

\title{Enhancing Tropical Cyclone Path Forecasting with an Improved Transformer Network\

}
\makeatletter
\newcommand{\linebreakand}{%
  \end{@IEEEauthorhalign}
  \hfill\mbox{}\par
  \mbox{}\hfill\begin{@IEEEauthorhalign}
}
\makeatother

\author{
  \IEEEauthorblockN{1\textsuperscript{st} Nguyen Van Thanh}
  \IEEEauthorblockA{\textit{Faculty of Electronics and Telecommunications} \\
    \textit{University of Engineering and Technology, VNU} \\
    Hanoi, Vietnam \\
    felicenguyen2001@gmail.com}
  \and 
  \IEEEauthorblockN{2\textsuperscript{nd} Nguyen Dang Huynh}
  \IEEEauthorblockA{\textit{Faculty of Information Technology} \\
    \textit{University of Engineering and Technology, VNU} \\
    Hanoi, Vietnam \\
    nguyendanghuynh1804@gmail.com}
  \linebreakand 
  \IEEEauthorblockN{3\textsuperscript{rd} Nguyen Ngoc Tan}
  \IEEEauthorblockA{\textit{Faculty of Electronics and Telecommunications} \\
    \textit{University of Engineering and Technology, VNU} \\
    Hanoi, Vietnam \\
    tan.nguyen@vnu.edu.vn}
  \and 
  \IEEEauthorblockN{4\textsuperscript{th} Nguyen Thai Minh}
  \IEEEauthorblockA{\textit{Faculty of Electronics and Telecommunications} \\
    \textit{University of Engineering and Technology, VNU} \\
    Hanoi, Vietnam \\
    minhnt@hnmu.edu.vn}
  \linebreakand
  \IEEEauthorblockN{5\textsuperscript{th} Nguyen Nam Hoang}
  \IEEEauthorblockA{\textit{Faculty of Electronics and Telecommunications} \\
    \textit{University of Engineering and Technology, VNU} \\
    Hanoi, Vietnam \\
    hoangnn@vnu.eduvn}
}

\maketitle

\begin{abstract}
A storm is a type of extreme weather. Therefore, forecasting the path of a storm is extremely important for protecting human life and property. However, storm forecasting is very challenging because storm trajectories frequently change. In this study, we propose an improved deep learning method using a Transformer network to predict the movement trajectory of a storm over the next 6 hours. The storm data used to train the model was obtained from the National Oceanic and Atmospheric Administration (NOAA) [1]. Simulation results show that the proposed method is more accurate than traditional methods. Moreover, the proposed method is faster and more cost-effective.
\end{abstract}


\section{Introduction}

Tropical cyclones are considered extreme weather phenomena. Major storms (with wind speeds of 63 km/h or higher) can cause significant damage to life and property for people in coastal areas worldwide \cite{ref2}. Each year, around 40 to 50 tropical depressions (formed near the equator) develop into storms globally. Accurate and timely storm path forecasting enables authorities to take appropriate preventive measures, minimizing damage caused by storms. However, due to the complexity of storm movements and global climate change, predicting storm trajectories faces many difficulties.

Storm path forecasting technology has made remarkable progress in recent years. However, most current forecasting methods are essentially statistical \cite{ref3}. Specifically, the U.S. National Hurricane Center (NOAA) currently uses several forecasting methods such as:

\begin{itemize}
    \item Global Forecast System (GFS) \cite{ref4} and the Hurricane Weather Research and Forecasting model (HWRF);
    \item Geostationary Operational Environmental Satellite (GOES) system \cite{ref5};
    \item Weather sensor network over land and sea.
\end{itemize}

First, the GFS and HWRF models compute and predict storm movements using algorithms and equations based on observational data and current weather conditions. However, the accuracy of GFS relies on the quality and precision of input data such as wind intensity, pressure, and weather information. This poses challenges in forecasting unpredictable factors such as rapid storm development and transformation.

The second system, GOES, collects continuous data on storm trajectories and intensities. By monitoring storm changes in real time, NOAA can provide updated information and forecasts. However, this depends on the satellite’s data collection and transmission capabilities, which can be affected by adverse weather or technical issues. Limited satellite coverage over remote oceanic areas also presents challenges.

Finally, NOAA uses a network of weather sensors to collect data on factors such as pressure, wind, temperature, and humidity. This method requires deploying and maintaining a complex system of sensors and data collection devices over a wide area, which can be difficult in remote or isolated oceanic regions.

The above methods are limited by the complexity and nonlinearity of atmospheric systems and the computational errors in solving complex equations.

Recently, some storm forecasting approaches based on deep learning have emerged, such as Recurrent Neural Networks (RNNs), Long Short-Term Memory (LSTM) networks \cite{ref6}, and Gated Recurrent Units (GRU), which are capable of storing temporal information effectively. Previously, Convolutional Neural Networks (CNNs) have been applied due to their ability to detect and learn spatial features in data, which makes them suitable for predicting storm trajectories based on the spatial characteristics of storm paths \cite{ref7}. However, these methods still have several limitations, such as difficulty in handling long-term storm trajectories. The models often only consider spatial features at each individual time step without capturing spatial interactions between data points along the trajectory. This is particularly problematic when the storm trajectory is large in scale or when multiple features and data sources need to be processed simultaneously. Traditional deep learning models like RNNs, GRUs, or CNNs may struggle to scale effectively for such complex tasks and large datasets \cite{ref9}.

The Transformer model, with its ability to learn nonlinear relationships among meteorological variables such as humidity, pressure, location, etc., and its capacity to handle sequences of varying lengths, holds great potential for improving the accuracy of storm trajectory prediction. Therefore, in this paper, we propose using the Transformer model to address the limitations of the aforementioned methods.

The structure of this paper is organized as follows:

The architecture and operational flow of the original Transformer model and our proposed version are presented in Sections~\ref{sec:transformer_model} and III, respectively. Finally, in Section IV, we compare the storm trajectory prediction results using the proposed Transformer model with those of other deep learning models such as LSTM, as well as with forecasting methods currently used by the U.S. National Weather Service.

\section{Transformer Model}
\label{sec:transformer_model}

The Transformer model is a neural network architecture for processing sequential data, introduced by Vaswani et al. in 2017 \cite{ref10}. It utilizes the Multi-Head Attention mechanism and a Feedforward Neural Network, and has demonstrated high effectiveness in various natural language processing tasks. Therefore, in this study, we apply the Transformer model for tropical cyclone trajectory forecasting to leverage its advantages—particularly its ability to learn nonlinear relationships between meteorological factors such as humidity, pressure, temperature, wind speed, etc., through the Attention layers.

This capability enables the model to capture correlations among different weather variables and improve forecasting accuracy. Moreover, the Transformer model offers flexibility in choosing the appropriate model size to match the complexity of input data and computational requirements. As shown in Figure~\ref{fig:transformer_architecture}, the Transformer architecture consists of two main components: the \textbf{Encoder} and the \textbf{Decoder}. Both the encoder and decoder are composed of multiple layers, each of which contains an \textit{Attention} layer and a \textit{Feedforward} layer. The encoder is used to extract features from the input data, while the decoder utilizes these features to generate predictions \cite{ref11}.

\begin{figure}[h]
    \centering
    \includegraphics[width=0.9\linewidth]{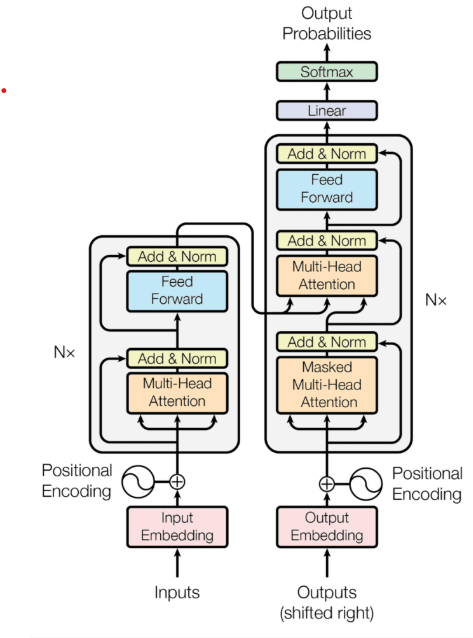}
    \caption{Overview of the original Transformer architecture, consisting of encoder and decoder components.}
    \label{fig:transformer_architecture}
\end{figure}

Several recent studies have also applied improved Transformer models to tasks such as drought prediction \cite{ref12}, and tracking and forecasting the intensity of tropical cyclones \cite{ref13}. In \cite{ref12}, the authors proposed a deep Transformer model with four encoder and decoder layers. The results showed that the Transformer achieved better long-term forecasting performance than traditional LSTM models. The study in \cite{ref13}, published in July 2023, proposed a Transformer network for simultaneously tracking storm trajectories and forecasting storm intensities. In this work, the authors used typhoon data from China for experimentation, and the results outperformed those of RNN-based models.

\section{Proposed Transformer Model Combined with Coordinate Grid}

Based on the analysis of the Transformer architecture above, in this section, we propose a model that combines the Transformer model combined with a coordinate grid to enhance accuracy and reduce the complexity of the method, as presented below.

\begin{figure}[h]
    \centering
    \includegraphics[width=0.9\linewidth]{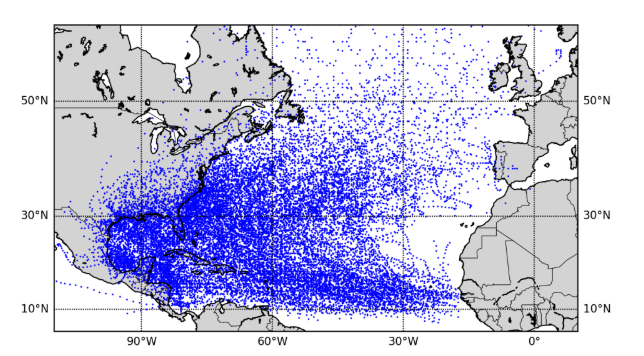}
    \caption{Storm occurrence points from 1944 to 2022}
    \label{fig:transformer_architecture}
\end{figure}

\begin{figure}[h]
    \centering
    \includegraphics[width=0.9\linewidth]{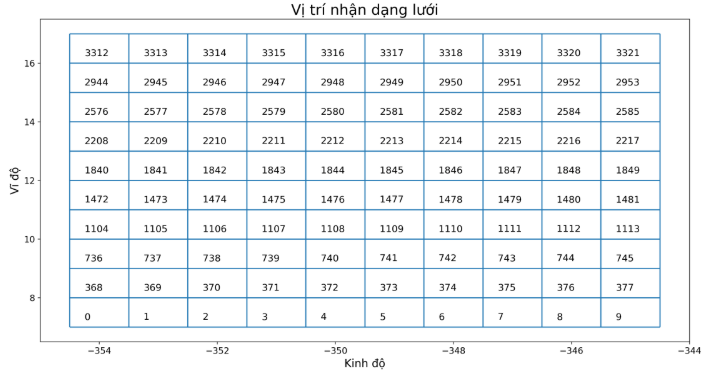}
    \caption{A coordinate grid with a resolution of 1° longitude × 1° latitude.}
    \label{fig:transformer_architecture}
\end{figure}

\subsection*{A. Grid Model}

Figure~\ref{fig:grid_points} shows the longitude and latitude points of tropical cyclones recorded by the Unisys Weather Dataset used for training and testing the model \cite{ref14}. This figure illustrates the typical movement of storm trajectories under given atmospheric conditions. A refined grid is placed over the longitude and latitude points to reduce truncation errors while allowing the model to capture large-scale patterns and represent small-scale phenomena more precisely. This setup is ideal for enabling the Transformer to optimally capture the complexity of storm trajectories.

In this paper, we propose training the Transformer model in combination with the coordinate grid model as presented in \cite{ref15}. Specifically, the Transformer learns the movement of a storm trajectory from one grid cell to another. The grid model illustrated in Figure~\ref{fig:grid_model} contains a total of 23,533 identifiable grid points.

\subsection*{B. Dataset}

We use data provided by the U.S. National Hurricane Center (NOAA) \cite{ref1}. The dataset includes information about tropical cyclones in the Atlantic basin from 1852 to 2022. Each data point contains the location of the storm's center every 6 hours (latitude and longitude), wind speed (in knots), and minimum central pressure. After removing years with faulty or missing data, the final dataset includes storms from 1944 to 2022, totaling 982 storms with 22,545 data records (each 6 hours apart).

From the storm's longitude and latitude data, we compute additional features such as distance and direction of movement to support trajectory prediction. The distance (in miles) between the current location and the next forecasted location (6 hours ahead) is calculated using the \texttt{Geopy} library, which allows us to measure the great-circle distance between two points given their coordinates.

Similarly, the direction of storm movement is computed using the angle $\beta$ between two consecutive positions as follows \cite{ref16}:

\begin{equation}
\angle \beta = \arctan \left( \frac{\sin(\Delta \lambda) \cos(\phi_2)}{\cos(\phi_1) \sin(\phi_2) - \sin(\phi_1) \cos(\phi_2) \cos(\Delta \lambda)} \right)
\end{equation}

where $\angle \beta \in [0, 360]$. If $\beta$ is negative, it is added with 360 to ensure a positive value. $(\lambda_1, \phi_1)$ and $(\lambda_2, \phi_2)$ represent the longitude and latitude of the storm at the current and next 6-hour step, respectively, and $\Delta \lambda = \lambda_2 - \lambda_1$.

We also compute the grid identifier (\texttt{gridID}) corresponding to the storm’s location on the coordinate grid as:

\begin{equation}
\text{gridID} = \left\lfloor \lambda - \lambda_{\text{min}} \right\rfloor \Phi + \left\lfloor \phi - \phi_{\text{min}} \right\rfloor
\end{equation}

where $\lfloor x \rfloor$ denotes the floor function, $[x]$ denotes the nearest integer function, and $\Phi = \phi_{\text{max}} - \phi_{\text{min}}$.

\subsection*{C. Data Processing}

The Transformer model requires sequential input data. However, storm trajectories vary in length. To address this, we apply zero-padding at the end of each storm sequence to standardize their lengths. The longest storm sequence in our dataset contains 96 time steps, so we pad all sequences to a fixed length of 100.

As a result, the training dataset contains 98,200 samples, each with 5 features: wind speed, pressure, distance, direction of movement, and grid identifier.

\begin{figure}[h]
    \centering
    \includegraphics[width=0.9\linewidth]{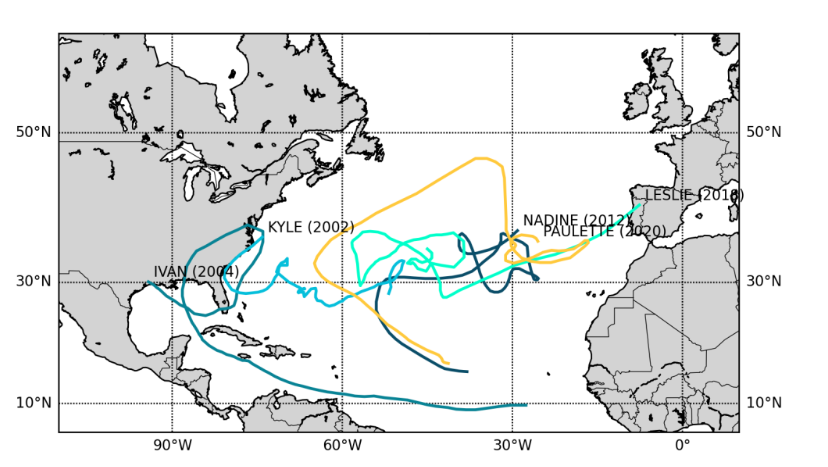}
    \caption{Trajectories of the 5 longest-lasting storms}
    \label{fig:transformer_architecture}
\end{figure}

\begin{figure}[h]
    \centering
    \includegraphics[width=0.5\linewidth]{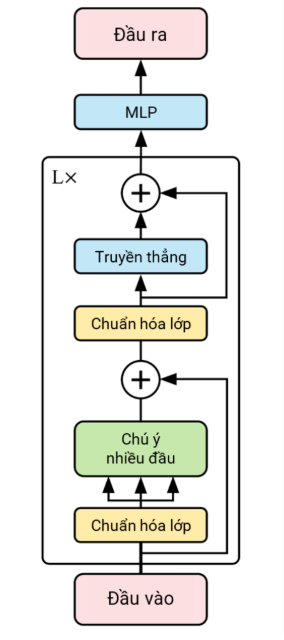}
    \caption{Proposed Transformer model architecture}
    \label{fig:transformer_architecture}
\end{figure}

Next, the data is segmented, with each segment containing 13 elements. The first 12 elements are used as input, and the grid identifier of the 13\textsuperscript{th} element is used as the prediction label. Finally, the dataset is divided into training and testing sets at a ratio of 85:15 to evaluate the model's performance.

\subsection*{D. Proposed Transformer Model Architecture}

Figure~\ref{fig:proposed_transformer} illustrates the proposed Transformer model architecture in this paper. Once the data has been converted into sequential form, each sequence contains 12 time steps, and each step is represented as a 5-dimensional vector. Within a sequence, the Attention layer enables each position to observe other positions, thereby re-encoding the current input by focusing on more important parts of the sequence. In particular, this attention mechanism is enhanced through multi-head attention, allowing each position to attend to multiple relevant features simultaneously.

These transformed elements are then passed through a Feedforward layer with the Gaussian Error Linear Unit (GELU) activation function \cite{ref17}, which improves the representation of vectors for the following layers.

The data is processed through three encoder layers to obtain better feature representations. Finally, the outputs are fed into a Multi-Layer Perceptron (MLP) output layer with two sublayers: one with 12 neurons and the final one with a single neuron. The activation functions used are ReLU for the hidden layer and tanh for the output layer. The last neuron represents the predicted grid identifier.

The use of the hyperbolic tangent function (tanh) in the final activation, instead of ReLU or Sigmoid, enables the model to output values in the range \([-1, 1]\), allowing more effective directional prediction of storm movements \cite{ref15}.

Figure~\ref{fig:loss_plot} and Figure~\ref{fig:accuracy_plot} illustrate the training and testing loss and accuracy across epochs, respectively.

\section{Results and Evaluation}

The proposed Transformer model was trained on a Google Colab environment with a Tesla T4 GPU. Training lasted for approximately 6 minutes with 100 epochs to minimize loss. The model was trained using the Mean Squared Error (MSE) loss function, Adam optimizer, and evaluated using accuracy as a metric.

After training, the model achieved an MSE of 0.0086 on the test set, with an accuracy of 0.78, as shown in Table~\ref{tab:results}.

Figures~\ref{fig:ivan_prediction} and~\ref{fig:delta_prediction} illustrate successful storm trajectory predictions by the proposed Transformer model for Hurricane Ivan (2004) and Hurricane Delta (2020), respectively.

\begin{figure}[h]
    \centering
    \includegraphics[width=0.9\linewidth]{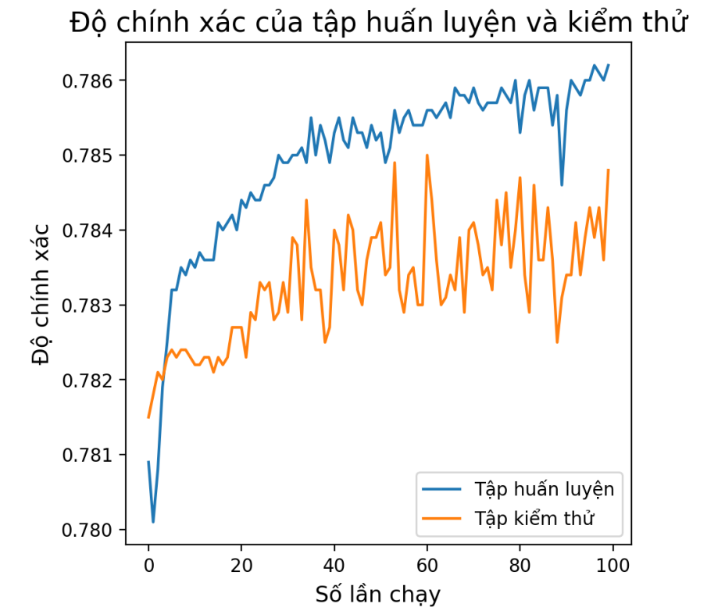}
    \caption{Chart showing the accuracy of the training and testing sets after each run}
    \label{fig:accuracy}
\end{figure}

\begin{figure}[h]
    \centering
    \includegraphics[width=0.9\linewidth]{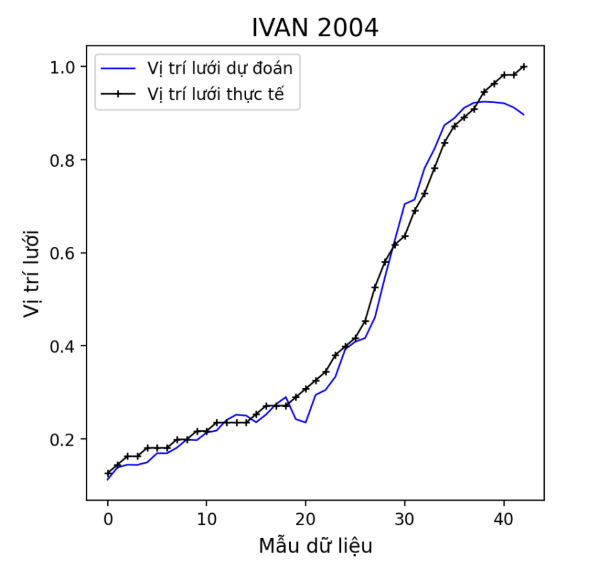}
    \caption{Predicted trajectory of Hurricane IVAN (2004)}
    \label{fig:ivan}
\end{figure}

\begin{figure}[h]
    \centering
    \includegraphics[width=0.9\linewidth]{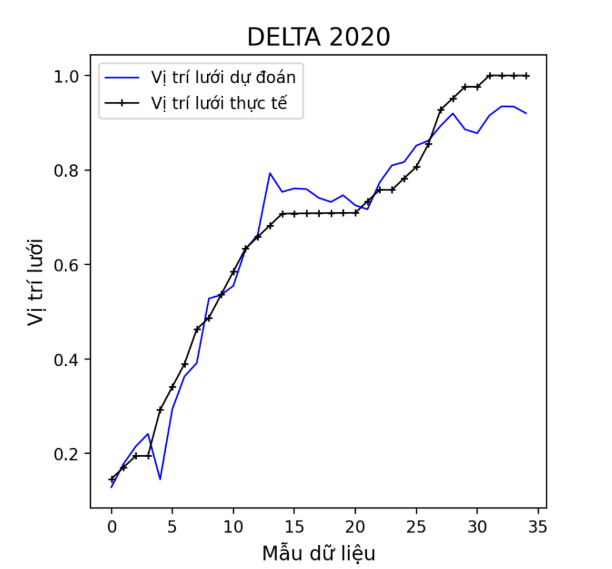}
    \caption{Predicted trajectory of Hurricane DELTA (2020)}
    \label{fig:delta}
\end{figure}

The model predicts storm movement in 6-hour intervals. As shown in the figures, the predicted grid locations closely match the actual storm paths, demonstrating the model's accuracy.

The performance metrics of the National Hurricane Center (NHC) are reported as part of the Government Performance and Results Act (GPRA) of 1993 \cite{ref14}. The current NHC metrics track the average annual forecast errors for storm position and intensity within the Atlantic basin over a 48-hour period for all tropical cyclones. Although the NHC provides forecasts at intervals from 12 to 120 hours, the 48-hour forecast is particularly important for emergency managers and preparation efforts. Due to natural variability in storm characteristics, annual errors may fluctuate significantly from year to year.

\begin{table}[h]
\centering
\caption{Comparison of Accuracy between LSTM Model and Proposed Transformer Model}
\label{tab:results}
\begin{tabular}{|l|c|c|}
\hline
\textbf{Model} & \textbf{MSE} & \textbf{Accuracy} \\
\hline
LSTM Model \cite{ref6} & 0.0160 & 0.685 \\
Proposed Transformer Model & 0.0086 & 0.783 \\
\hline
\end{tabular}
\end{table}

Figure~\ref{fig:gpra_comparison} shows that both the proposed Transformer-based forecasting method and NOAA's method outperform the minimum accuracy requirements set by GPRA. Furthermore, the proposed Transformer model using the coordinate grid provides better accuracy than existing NHC methods.

Additionally, the statistical-dynamic models currently used by the NHC often take several hours to produce a single forecast, utilizing some of the world’s most advanced supercomputers. In contrast, deep learning models such as the proposed Transformer can deliver accurate forecasts significantly faster.

However, the proposed Transformer model still faces several challenges in storm trajectory prediction. For instance, the model may suffer from overfitting if not properly regulated or if the dataset is imbalanced. This can lead to failure in trajectory forecasting. Moreover, the Transformer architecture is computationally intensive and complex, especially when trained on large datasets. This presents difficulties in deploying the model on resource-constrained devices.

\begin{figure}[h]
    \centering
    \includegraphics[width=0.9\linewidth]{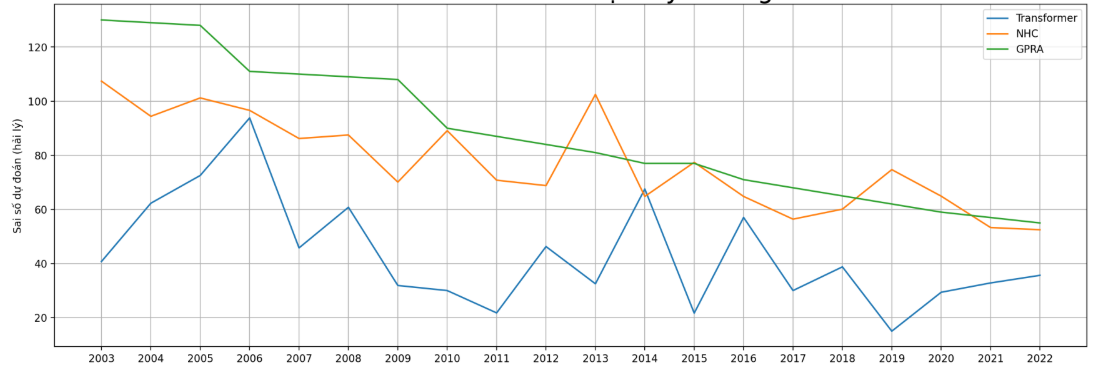}
    \caption{Comparison of forecast accuracy between NOAA, GPRA standards, and the proposed Transformer model.}
    \label{fig:gpra_comparison}
\end{figure}

\section{Conclusion}

In this study, we implemented a method for forecasting storm trajectories over continuous 6-hour intervals using a Transformer model combined with a grid-based coordinate mapping system. The model successfully predicted storm paths for the next 6 hours, achieving an accuracy of 0.783 and a mean squared error (MSE) of 0.0086. When compared with NOAA methods and GPRA benchmarks, our model demonstrated superior performance while also completing training more quickly.

\end{document}